\documentclass[10pt,twocolumn,letterpaper]{article}

\usepackage[accsupp]{axessibility} 
\usepackage{cvpr}
\usepackage[latin9]{inputenc}
\usepackage{amsmath,amssymb,amsfonts}
\usepackage{times}
\usepackage{graphicx}
\usepackage{epsfig}
\usepackage{multirow}
\usepackage{booktabs}
\usepackage{algorithmic}
\usepackage{siunitx}
\usepackage[export]{adjustbox}
\usepackage{subcaption}

\usepackage[norule]{footmisc}

\usepackage{float} 

\usepackage[pagebackref=true,breaklinks=true,bookmarks=false]{hyperref}

\graphicspath{{figures/}}
\DeclareGraphicsExtensions{.eps,.png,.pdf,.jpg,.jpeg,.JPG}

\begin{document}

\title{Vulnerability of Automatic Identity Recognition to Audio-Visual Deepfakes}

\author{Pavel Korshunov
\and
Haolin Chen
\and
Philip N. Garner
\and
S\'{e}bastien Marcel
\and
Idiap Research Institute,
Martigny, Switzerland.\\
{\tt\small \{pavel.korshunov,haolin.chen,phil.garner,sebastien.marcel\}@idiap.ch}
}

\maketitle
\thispagestyle{empty}

\begin{abstract}

The task of deepfakes detection is far from being solved by speech or vision researchers. Several publicly available databases of fake synthetic video and speech were built to aid the development of detection methods.  However, existing databases typically focus on visual or voice modalities and provide no proof that their deepfakes can in fact impersonate any real person. In this paper, we present the first realistic audio-visual database of deepfakes SWAN-DF, where lips and speech are well synchronized and video have high visual and audio qualities. We took the publicly available SWAN dataset of real videos with different identities to create audio-visual deepfakes using several models from DeepFaceLab and blending techniques for face swapping and HiFiVC, DiffVC, YourTTS, and FreeVC models for voice conversion. From the publicly available speech dataset LibriTTS, we also created a separate database of only audio deepfakes LibriTTS-DF using several latest text to speech methods: YourTTS, Adaspeech, and TorToiSe. We demonstrate the vulnerability of a state of the art speaker recognition system, such as ECAPA-TDNN-based model from SpeechBrain, to the synthetic voices. Similarly, we tested face recognition system based on the MobileFaceNet architecture to several variants of our visual deepfakes. The vulnerability assessment show that by tuning the existing pretrained deepfake models to specific identities, one can successfully spoof the face and speaker recognition systems in more than 90\% of the time and achieve a very realistic looking and sounding fake video of a given person.
\end{abstract}


\section{Introduction}
\label{sec:intro}

The original predictions that deepfakes would pose a significant danger to society\footnote{\scriptsize \url{https://edition.cnn.com/interactive/2019/01/business/pentagons-race-against-deepfakes/}} are turning out to be correct with more and more reports of money extortion with voice cloning\footnote{\scriptsize \url{https://www.nbc15.com/2023/04/10/ive-got-your-daughter-mom-warns-terrifying-ai-voice-cloning-scam-that-faked-kidnapping/}}, duping politicians using video deepfakes\footnote{\scriptsize \url{https://www.theguardian.com/world/2022/jun/25/european-leaders-deepfake-video-calls-mayor-of-kyiv-vitali-klitschko}}, or using deepfake pornography for revenge or harassment online\footnote{\scriptsize\url{https://edition.cnn.com/2023/02/16/tech/nonconsensual-deepfake-porn/index.html}}. 

%
%

Many databases with deepfake videos were created to help develop and train deepfake detection methods. One of the first freely available databases was DeepfakeTIMIT~\cite{Korshunov2019a}, followed by the FaceForensics database with deepfakes generated from \num{1000} Youtube videos~\cite{Verdoliva2018}, and which was later morphed into FaceForensics++ with more types of deepfakes and a separate set of original and deepfake videos provided by Google and Jigsaw~\cite{Roessler2019}. Several independent extensions of FaceForensics++ were also proposed, including the HifiFace~\cite{HifiFace2021} and DeeperForensics~\cite{Deeperforensics2020} datasets. Another \num{5000} videos-large database of deepfakes generated from Youtube videos is Celeb-DF v2~\cite{Celeb_DF_cvpr20}.  Facebook~\cite{DFDC2020} also created their own database with more than $100K$ deepfake videos, which was used in Deepfake Detection Challenge 2020 hosted by Kaggle\footnote{\scriptsize \url{https://www.kaggle.com/c/deepfake-detection-challenge}}. Mobio-DF\cite{Anubhav2021} is a dataset of $45K$ videos but with an unusually larger set of real videos compared to deepfakes. However, the largest database of deepfake videos to date is the Korean Deepfake (KoDF) dataset~\cite{KoDF2021} with about $175K$ fake videos.

One of the important issues with the most of the existing deepfake databases is that very little is known about the quality of the deepfake videos in terms of their ability to actually impersonate a targeted person. Besides a limited study of  face recognition vulnerability~\cite{Korshunov2019a}, the authors of datasets do not provide any justification of whether their deepfakes even look like a person, let alone a specific person. It means that without a verification of how fake  those deepfakes are, even a slight distortion of an original video (e.g., an applied color correction) may be considered as a deepfake. This actually happens, as we can observe some videos from Facebook~\cite{DFDC2020} dataset, which are labeled as being deepfake, to only have a moving patch of Gaussian noise as the only visible difference from the original real version. The lack of clear understanding of what constitutes a deepfake in a dataset leads to an over-fit problem, when detection algorithms, trained on such \textit{so called} deepfakes, may end up detecting distortions that are irrelevant to those manifested in the realistic and dangerous deepfakes.

\newcolumntype{M}[1]{>{\centering\arraybackslash}m{#1}} 

\begin{figure*}[p]
\setlength\tabcolsep{1pt} 
\centering
\begin{tabular}{@{} M{0.16\linewidth} M{0.16\linewidth} M{0.16\linewidth} M{0.16\linewidth} M{0.16\linewidth} M{0.16\linewidth} @{}}
{\small Original source} & {\small Original target} & {\small 160px, no correction} & {\small 160px, color adjusted} & {\small 256px, params tuned} & {\small 320px, params tuned}\\

\includegraphics[width=0.153\textwidth]{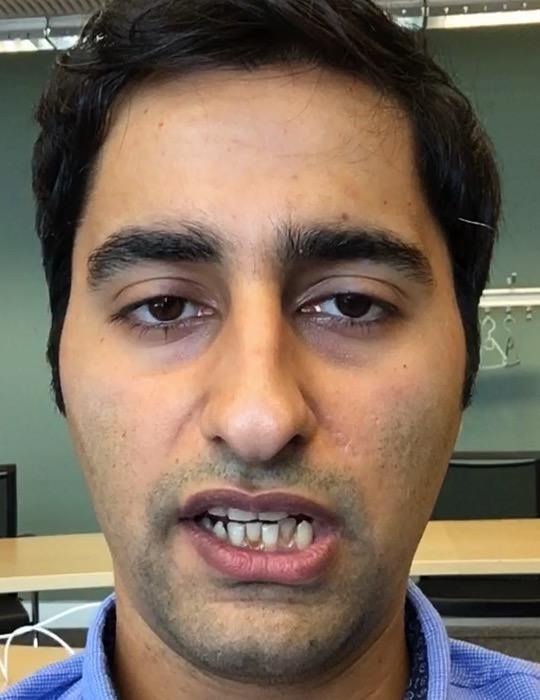}
& \includegraphics[width=0.153\textwidth]{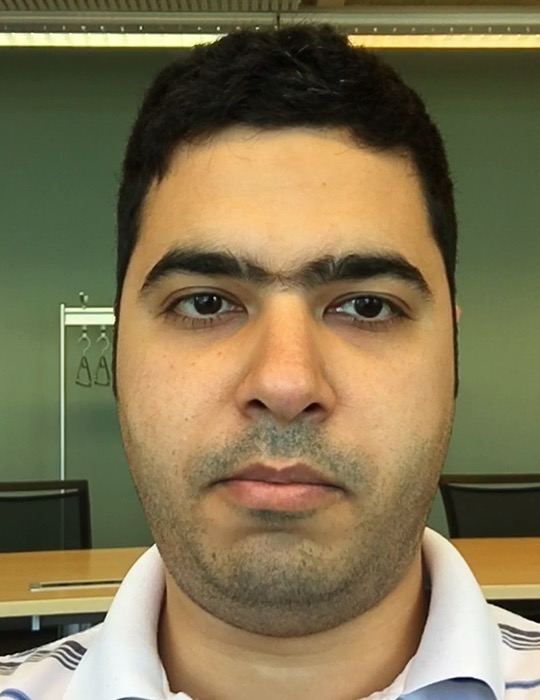}
& \includegraphics[width=0.153\textwidth]{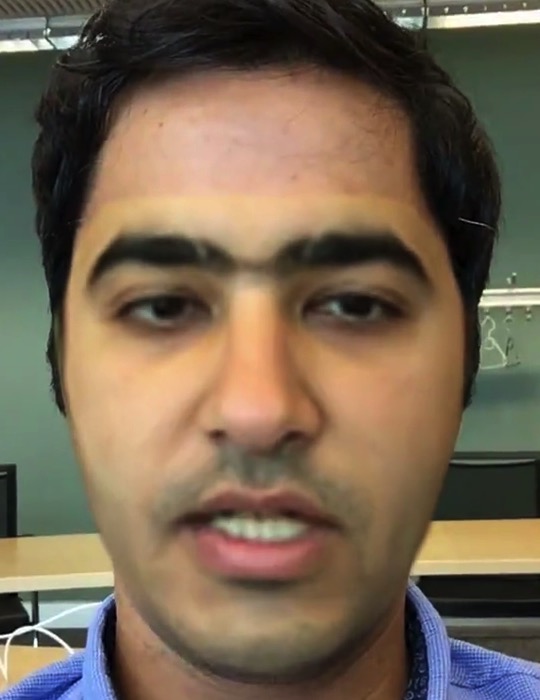}
& \includegraphics[width=0.153\textwidth]{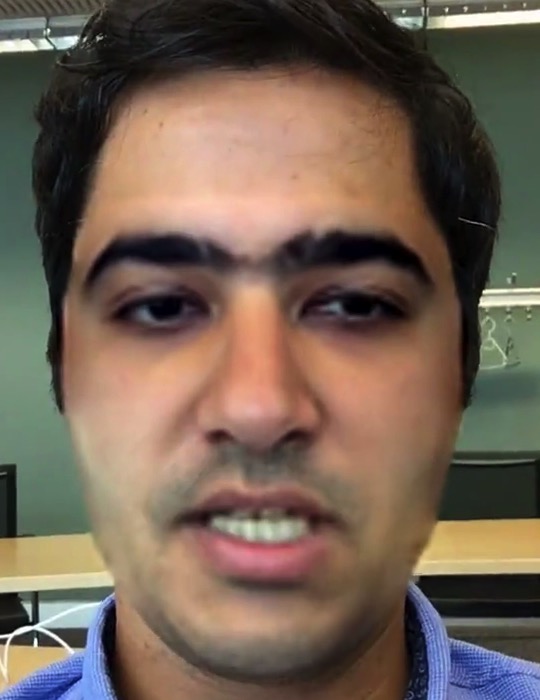}
& \includegraphics[width=0.153\textwidth]{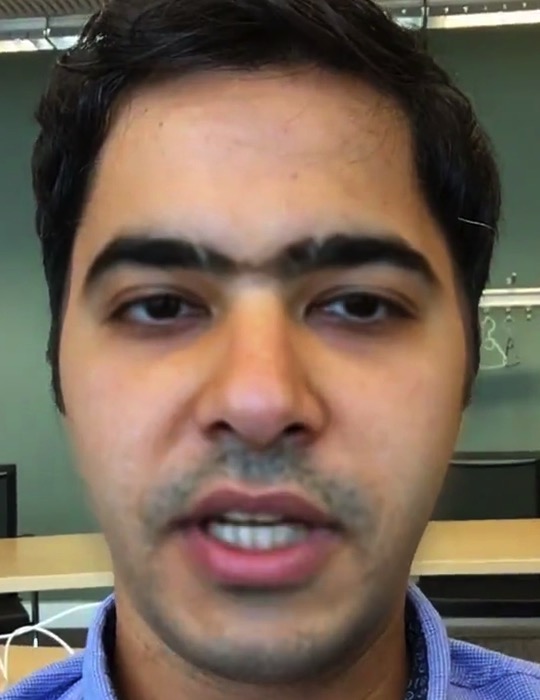}
& \includegraphics[width=0.153\textwidth]{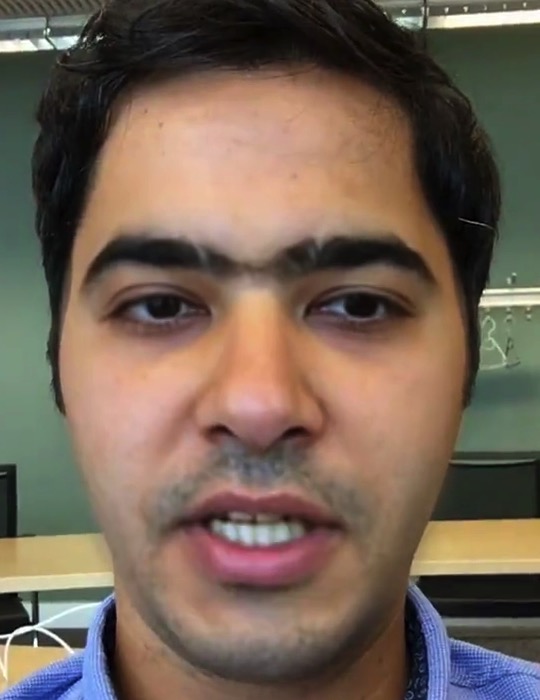}
\\

\includegraphics[width=0.153\textwidth]{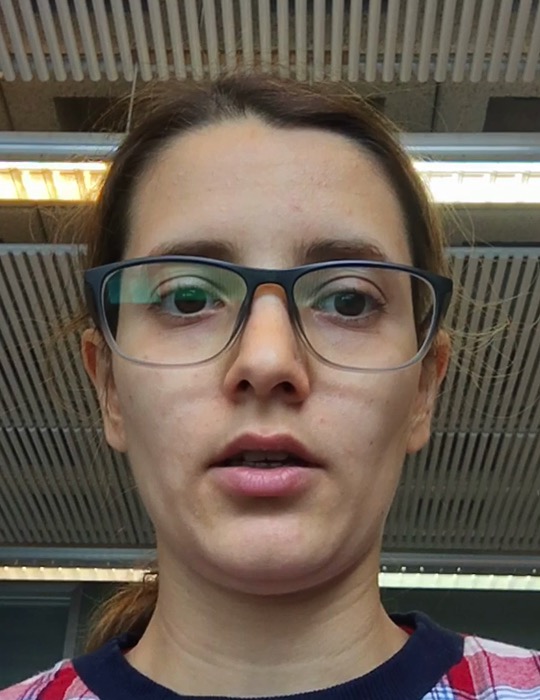}
& \includegraphics[width=0.153\textwidth]{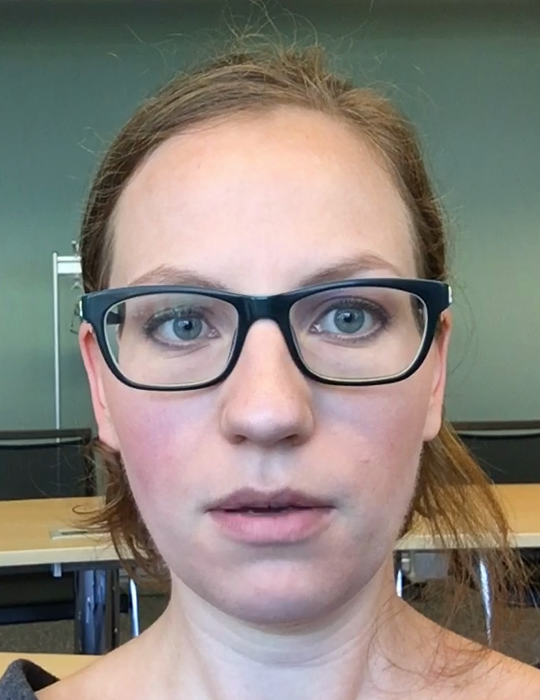}
& \includegraphics[width=0.153\textwidth]{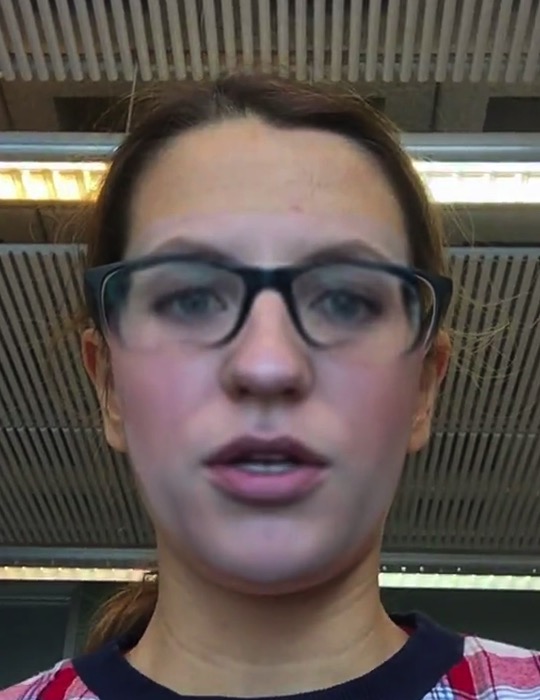}
& \includegraphics[width=0.153\textwidth]{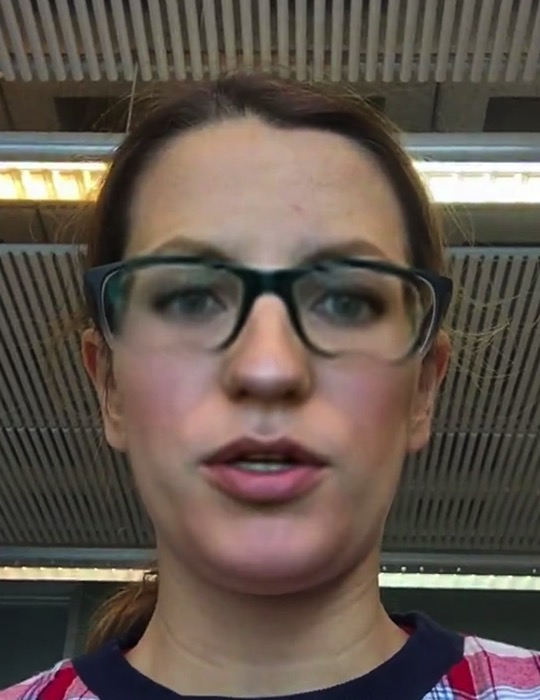}
& \includegraphics[width=0.153\textwidth]{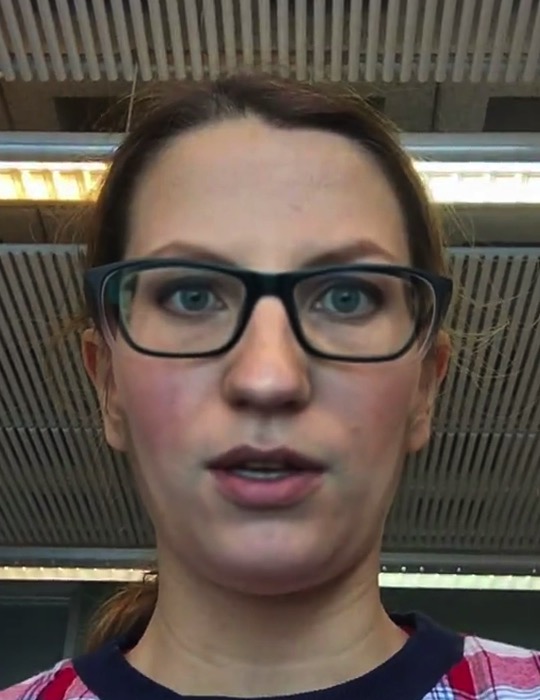}
& \includegraphics[width=0.153\textwidth]{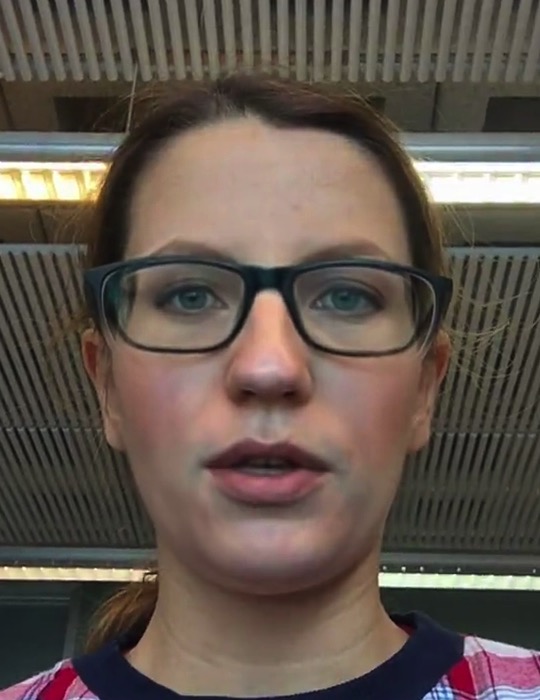}
\\

\includegraphics[width=0.153\textwidth]{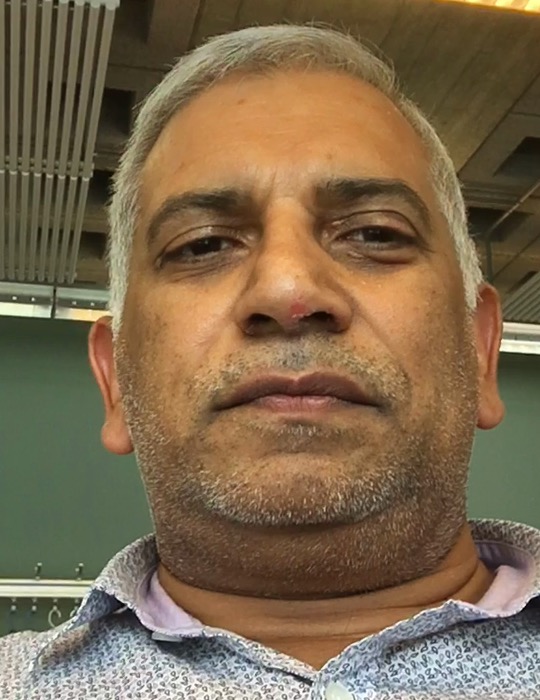}
& \includegraphics[width=0.153\textwidth]{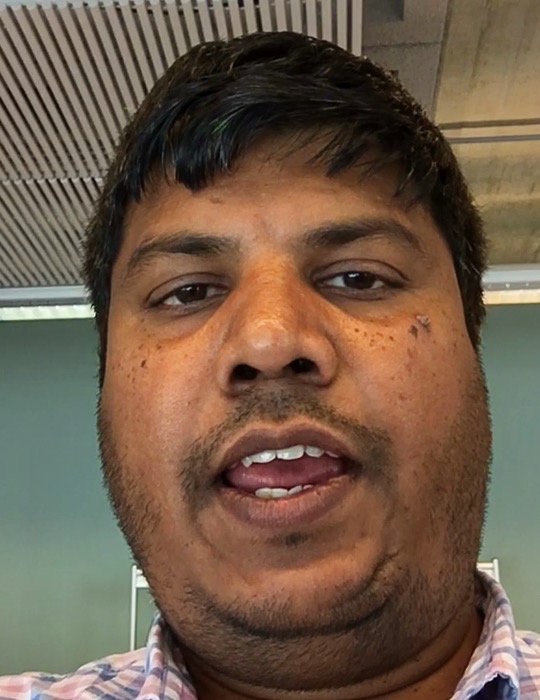}
& \includegraphics[width=0.153\textwidth]{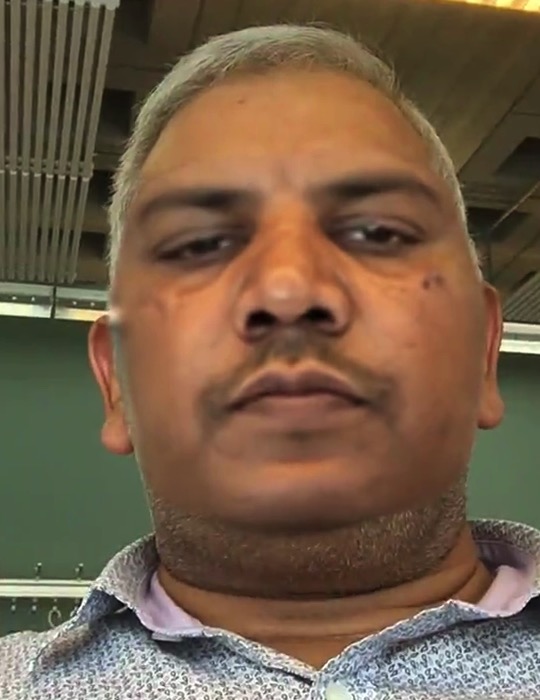}
& \includegraphics[width=0.153\textwidth]{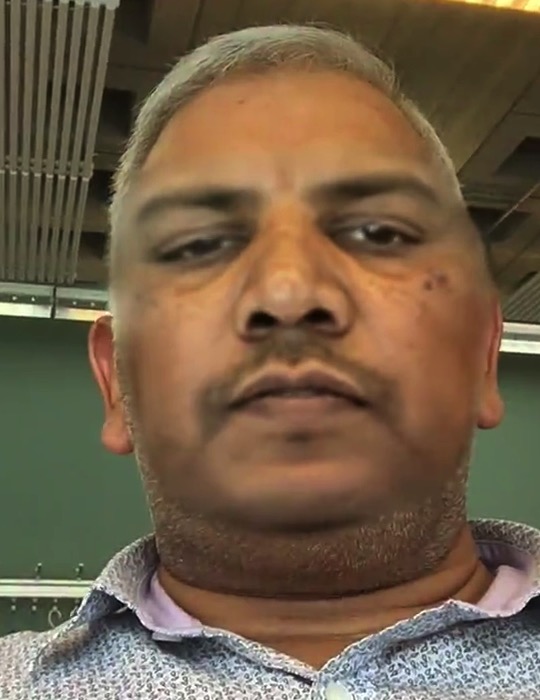}
& \includegraphics[width=0.153\textwidth]{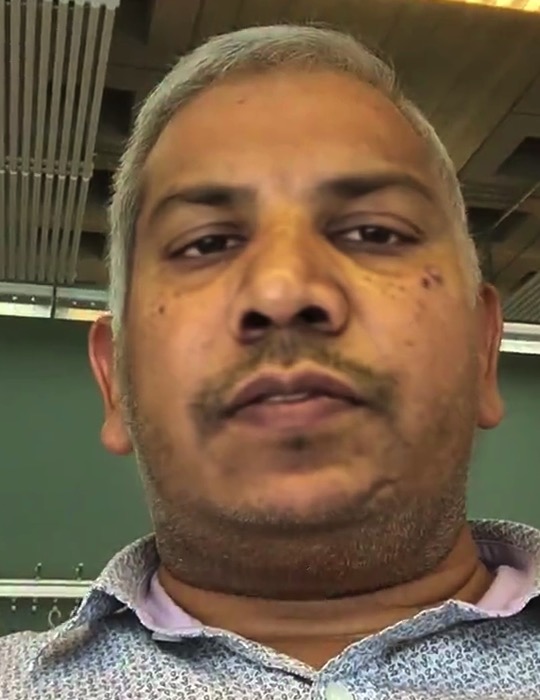}
& \includegraphics[width=0.153\textwidth]{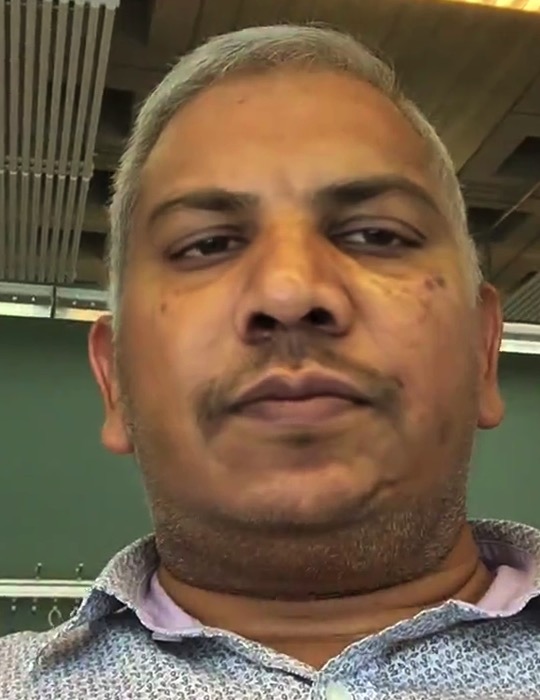}
\\

\includegraphics[width=0.153\textwidth]{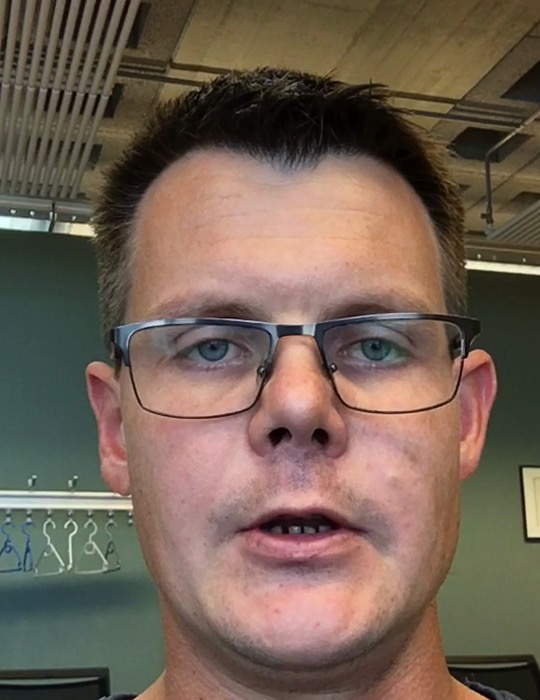}
& \includegraphics[width=0.153\textwidth]{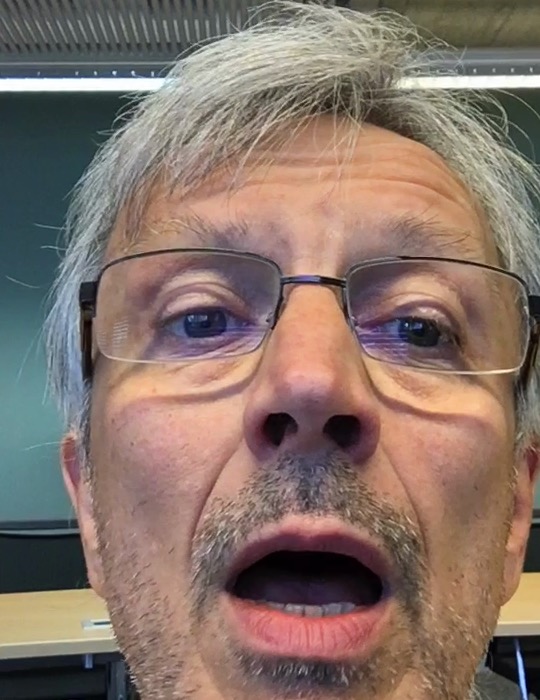}
& \includegraphics[width=0.153\textwidth]{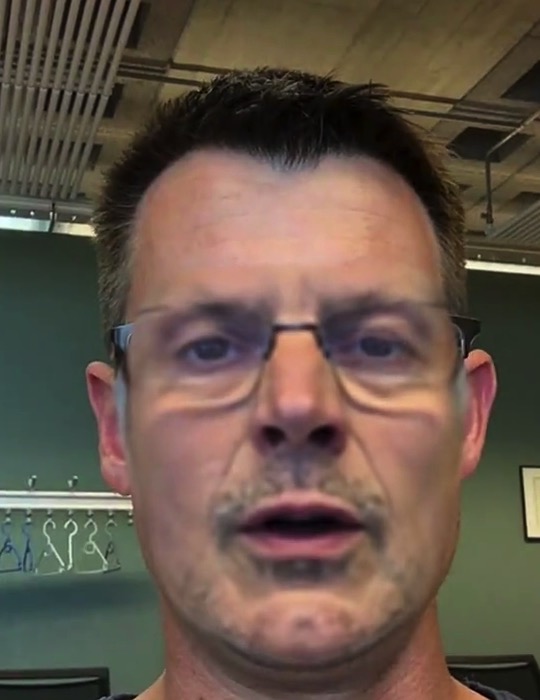}
& \includegraphics[width=0.153\textwidth]{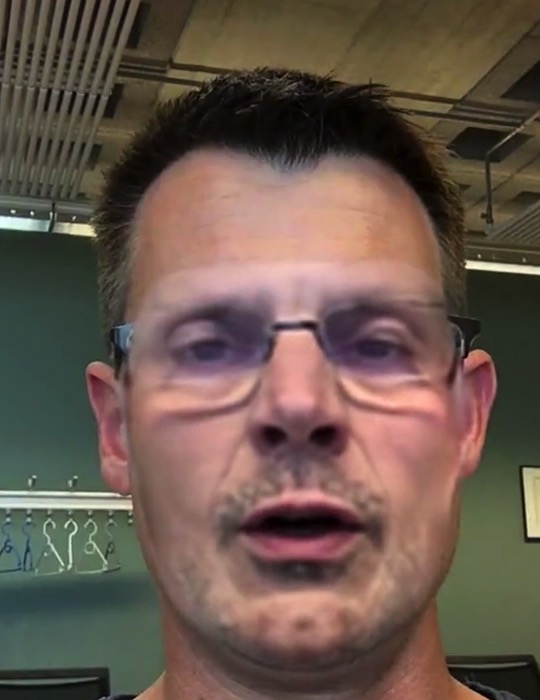}
& \includegraphics[width=0.153\textwidth]{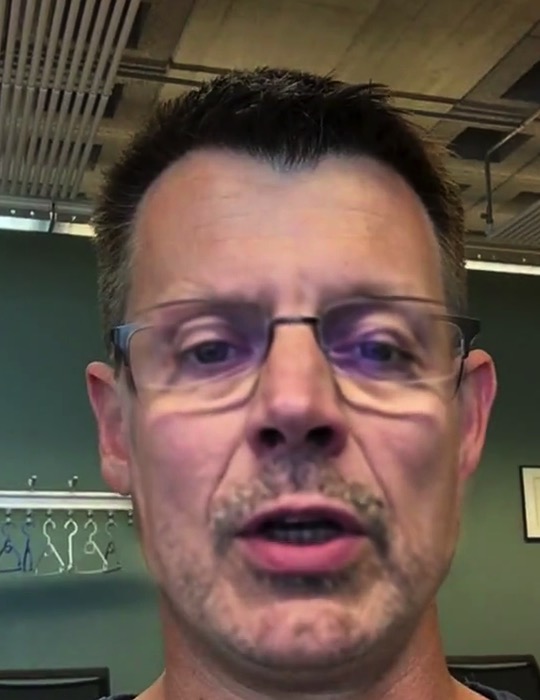}
& \includegraphics[width=0.153\textwidth]{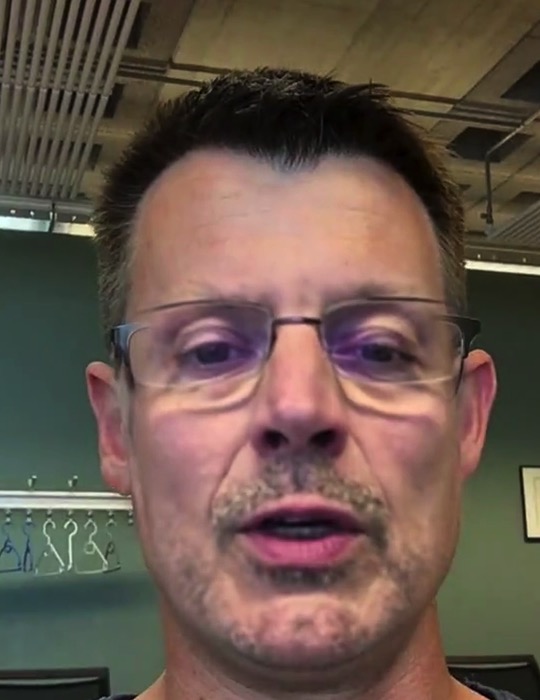}
\\

\includegraphics[width=0.153\textwidth]{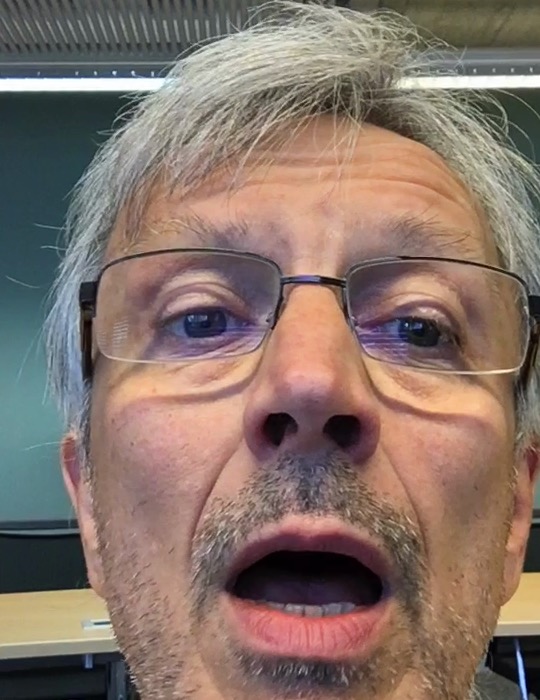}
& \includegraphics[width=0.153\textwidth]{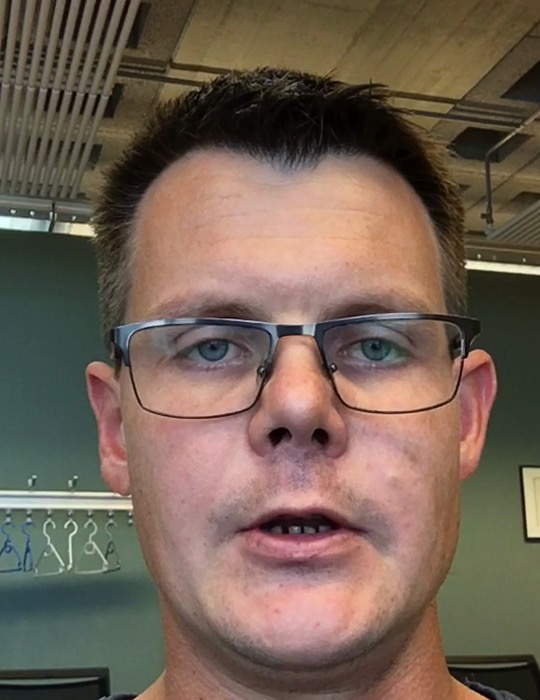}
& \includegraphics[width=0.153\textwidth]{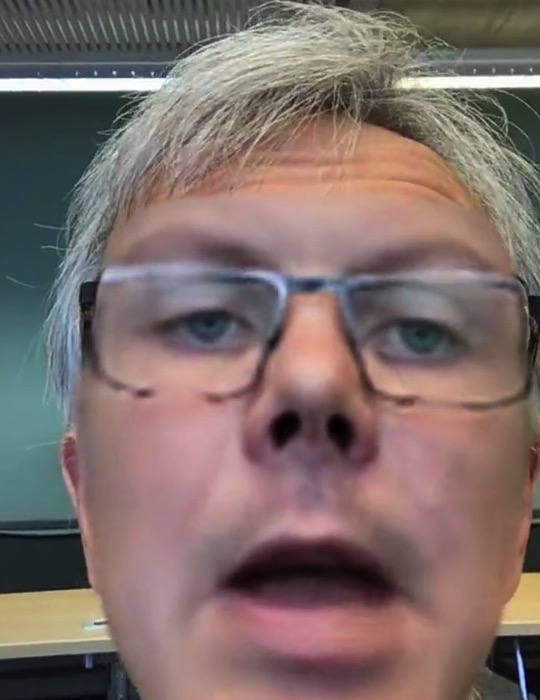}
& \includegraphics[width=0.153\textwidth]{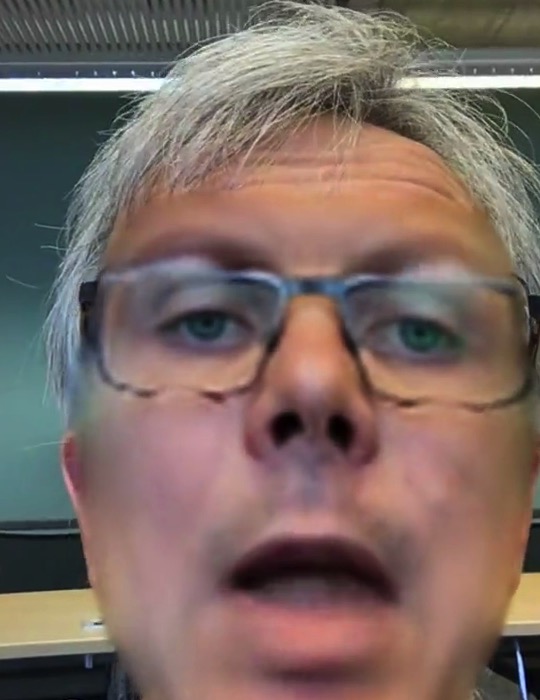}
& \includegraphics[width=0.153\textwidth]{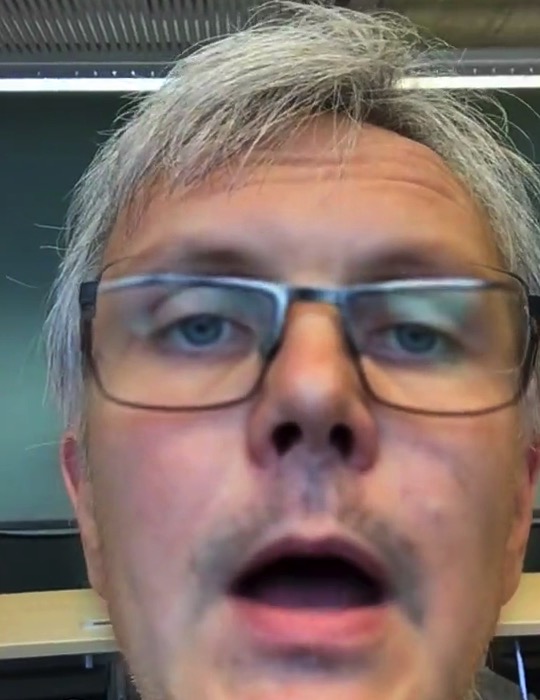}
& \includegraphics[width=0.153\textwidth]{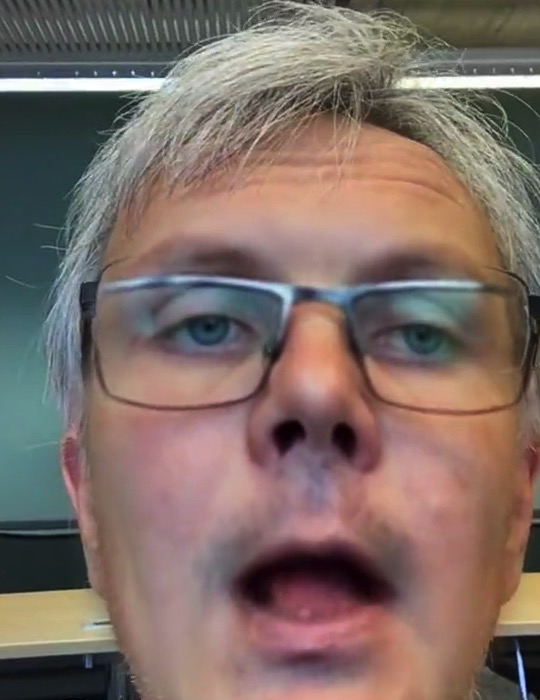}
\\

\includegraphics[width=0.153\textwidth]{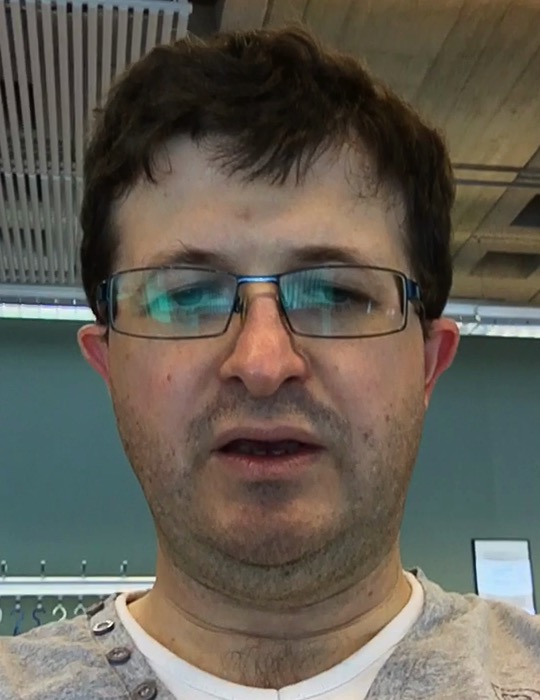}
& \includegraphics[width=0.153\textwidth]{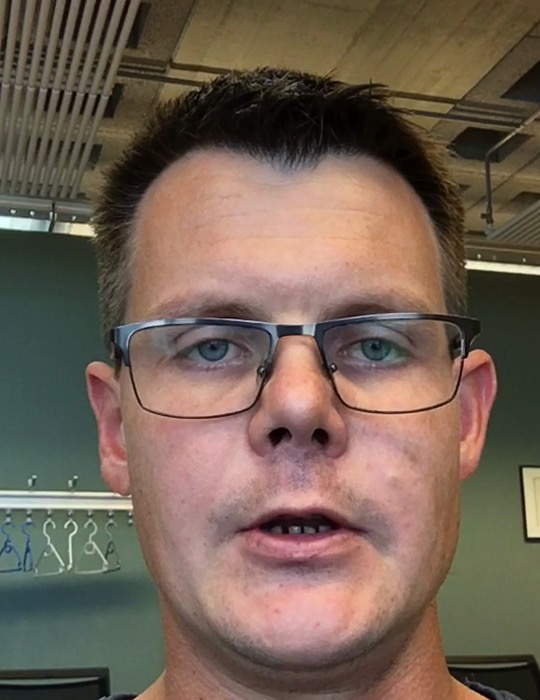}
& \includegraphics[width=0.153\textwidth]{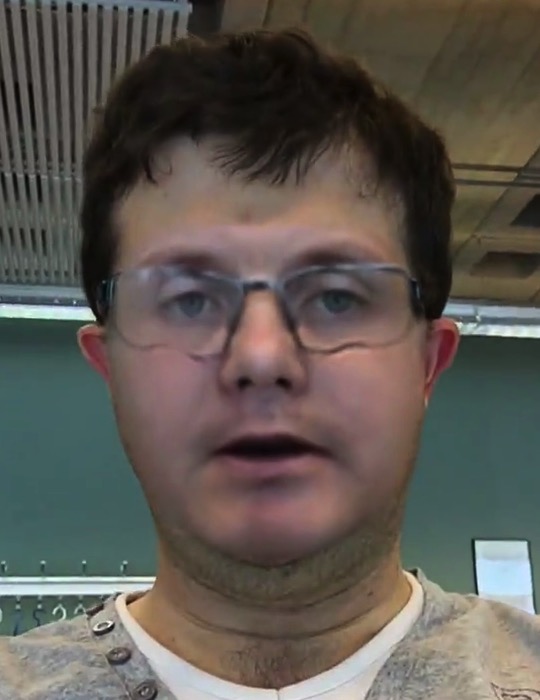}
& \includegraphics[width=0.153\textwidth]{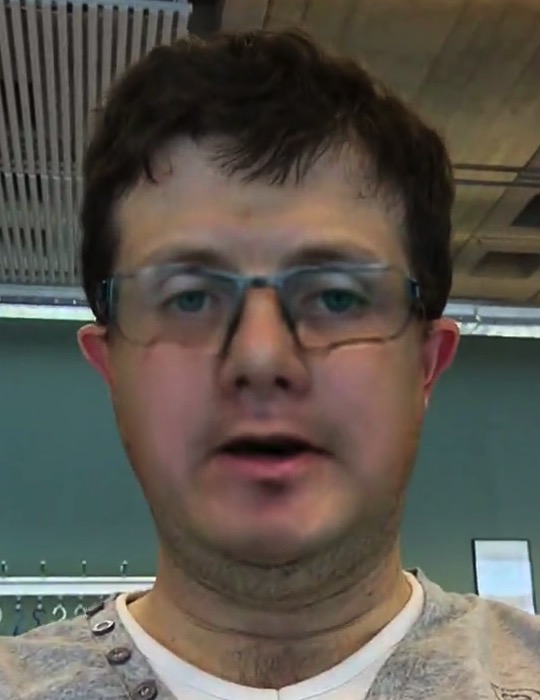}
& \includegraphics[width=0.153\textwidth]{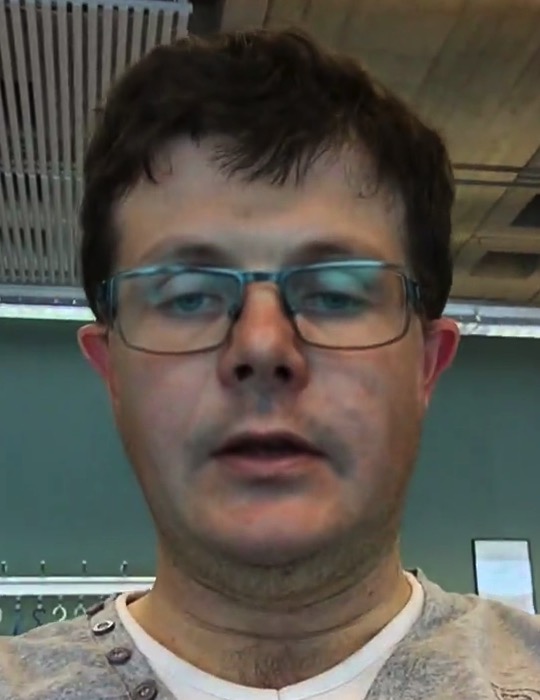}
& \includegraphics[width=0.153\textwidth]{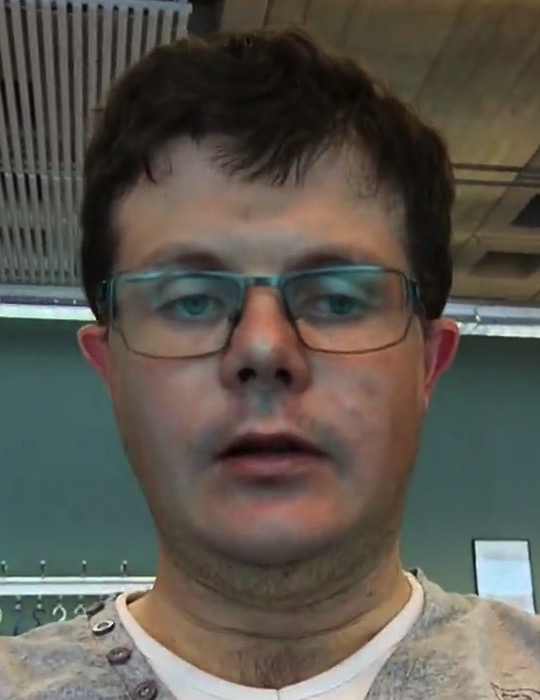}
\\
\subcaption{Original source}\label{subfig:a} & \subcaption{Original target}\label{subfig:b} & \subcaption{160px, no correction}\label{subfig:c} & \subcaption{160px, color adjusted}\label{subfig:d} & \subcaption{256px, params tuned}\label{subfig:e} & \subcaption{320px, params tuned}\label{subfig:f}\\
\end{tabular}

\caption{Examples of deepfakes (cropped to face) from the SWAN-DF database. Face of the `target' is placed into video of the `source'.}
\label{fig:swandf}
\end{figure*}

Another important issue, often overlooked by the authors of deepfake databases and the researchers who develop detection methods, is that the blending techniques used during face swapping process (or reenactment) arguably have as much effect on the accuracy of the detection as the generative adversarial networks (GANs) used to generated fake faces. This effect is well illustrated by the authors of~\cite{Efros2020}, who demonstrate that uncompressed GAN-generated images can be detected with a nearly 100\% accuracy, while the accuracy of the same detection methods degrade significantly on deepfake videos, where a similarly GAN-generated image is blended in and blurred into the compressed frame. It shows the possibility that many proposed deepfake detection methods may detect the distortions that come from blending and compression instead of the signatures that come from GANs. Arguably, this is why deepfake detection methods do not generalize well when they are tested on a database that used the same GAN architecture as the training videos but different blending methods~\cite{Generalizable2020,Korshunov2022,Korshunov2022b}.

Looking at the voice deepfakes, there are fewer databases that rely on neural networks for text to speech or voice conversion methods. The most notable database is the one used in ASVSpoof challenge in 2019 and 2021~\cite{yamagishi21_asvspoof}, which has a separate subset of audio deepfakes. These deepfakes were generated by several modern text to speech models to train and test detection methods but the methods generate either a voice of a single speaker or a limited pre-defined set of multiple speakers. No identity transfer was done by the methods used to generate these fakes and that is why the authors did not provide identity information for samples in the deepfake subset. 
WaveFake~\cite{WaveFake} is the latest dataset of deepfake speech that used public LJSpeech~\cite{ljspeech17} dataset of the single-speaker recordings as the original source to generate the fake samples. Until recently, most of the work in deep learning based methods that generate fake speech focused on producing the realistically sounding voice samples and little was done to preserve identity in that sample. However, recent advances in text to speech and voice conversion methods, allow to create datasets of truly deepfake speech which would preserve or transfer personal identity into the generated fake sample.

In this paper, we present the first high fidelity publicly available dataset  of realistic deepfakes SWAN-DF (see examples in Figure~\ref{fig:swandf}) where both faces and voices appear and sound like the target person. The SWAN-DF dataset is based on the public SWAN database~\cite{Swan2019} of real videos recorded in HD on iPhone and iPad Pro (in year $2019$). For $30$ pairs of people, we swapped faces and voices using several autoencoder-based face swapping models form the well-known open source repo DeepFaceLab\footnote{\scriptsize \url{https://github.com/iperov/DeepFaceLab}}~\cite{DeepFaceLab} and voice conversion (or voice cloning) methods, including zero-shot YourTTS~\cite{YourTTS} and various models from FreeVC~\cite{li2022freevc}. In addition to the audio-visual deepfake dataset, we also built LibriTTS-DF database (from a well-known LibriTTS~\cite{libritts} database), which contains fake speech samples for $39$ speakers generated with either text to speech methods that preserve intended identity, including our own adaptation of Adaspeech TTS model~\cite{chen2021adaspeech} and diffusion-based TorToiSe TTS\footnote{\scriptsize \url{https://github.com/neonbjb/tortoise-tts}} or YourTTS~\cite{YourTTS} zero-shot voice conversion approach. 

For video deepfakes, we also have put an effort into creating a large variety of different versions of generated videos in terms of models and blending techniques used. We have employed three different models with resolutions $160$px, $256$px, and $320$px, all pretrained on a large variety of faces by the contributors of DeepFaceLab. For each of $60$ people pairs from SWAN database, we have tuned each model type. And then, we also generated several version videos for each of this model, where we use different masking, color correction, and other blending parameters. In total, we generated more than $20$ deepfake variants for each video of each pair of people. These variations should allow to train and also test detection models that are invariant to the blending methods but instead focus on the distortions that are specific to deepfakes themselves, such as inconsistencies in accessories, issues with hair, eyes and teeth, geometrical facial distortions, etc.

To show how well different audio and visual deepfake generation methods preserve identity, we conducted an extensive vulnerability analysis using the ECAPA-TDNN-based state of the art speech recognition model from SpeechBrain\footnote{\scriptsize \url{https://speechbrain.github.io/}}, 
and MobileFaceNet~\cite{MobileFaceNet2018}, a popular pretrained PyTorch face recognition model\footnote{\scriptsize \url{https://github.com/foamliu/MobileFaceNet}}.

%
%
%
%

To allow researchers to use the database in a transparent manner and verify and reproduce our vulnerability evaluations, we provide the generated audio and video samples, list of files and splits into subsets, source code for vulnerability analysis and a jupyter notebook with complete results and graphs as an open-source Python package\footnote{\scriptsize Source code:  
\url{https://gitlab.idiap.ch/bob/paper.ijcb2023.av-deepfakes}}. Our Mobio-DF and LibriTTS-DF databases with examples of deepfake videos and voices can be found at the demo page\footnote{\scriptsize Database: \url{https://swan-df.github.io/}}.

\section{Related work}
\label{sec:related}

The generation approaches of synthetic faces can be split in four different categories i) completely synthetically generated images (identity is usually not preserved) using StyleGANs~\cite{stylegan,stylegan2}, ii) morphed images when faces of two people are morphed\footnote{\scriptsize \url{https://github.com/yaopang/FaceMorpher}}~\cite{Ferrara2014,Sarkar2022}, iii) face swapping based video deepfakes~\cite{Korshunova2017,DFDC2020,Roessler2019}, and iv) reenactment based deepfake videos~\cite{Deeperforensics2020,avatar2021,MegaPortraits2022}, which grew out of the idea of using a recurrent network to synthesize mouth texture directly from the voice~\cite{Shlizerman2017}. 

Methods for detecting visual fakes range from those based on simple visual or facial features~\cite{Zhang2017,Yang2019landmark,Agarwal2017,Li2018}, binary classifiers trained on fake images~\cite{Efros2020,Hulzebosch_2020_CVPR_Workshops} and videos~\cite{Roessler2019,Nguyen2019,Delp2020}, to the methods that try to generalize to new deepfake methods or various post-processing blending techniques~\cite{aneja2020generalized,Korshunov2022,Korshunov2022b}.

The state of the art in text to speech and voice conversion is represented by probabilistic generative models, particularly those based on diffusion~\cite{difftts,DBLP:journals/corr/abs-2111-14822/diffusiontransformer}, but also flow~\cite{fastspeech,fastspeech2,ddpm} or a combination or both~\cite{adaspeech4,guidedtts}.  Such techniques originated in the image processing or computer vision literature.  They are characterized by an iterative conversion from a simple distribution (that lends itself to sampling) to a complicated distribution representative of speech (but difficult to sample from).  At each iteration, a DNN is used to guide the conversion; in flow it is a transformation, in diffusion a denoizing process.

Similar to the work on detecting visual deepfakes, methods for detection of synthetic speech are also struggling with generalization to unseen attacks as is evident from the latest ASVspoof challenge and related work~\cite{yamagishi21_asvspoof,wang21_asvspoof}. Although the latest detection methods more and more rely on the end-to-end systems for feature extraction and modeling, the earlier work was often based on acoustic features~\cite{TODISCO2017516} and classical GMM-like modeling~\cite{Korshunov2017,KorshunovPAD2019}.

\begin{figure*}[tb]
\centering
\begin{subfigure}{0.33\textwidth}
	\includegraphics[width=\linewidth]{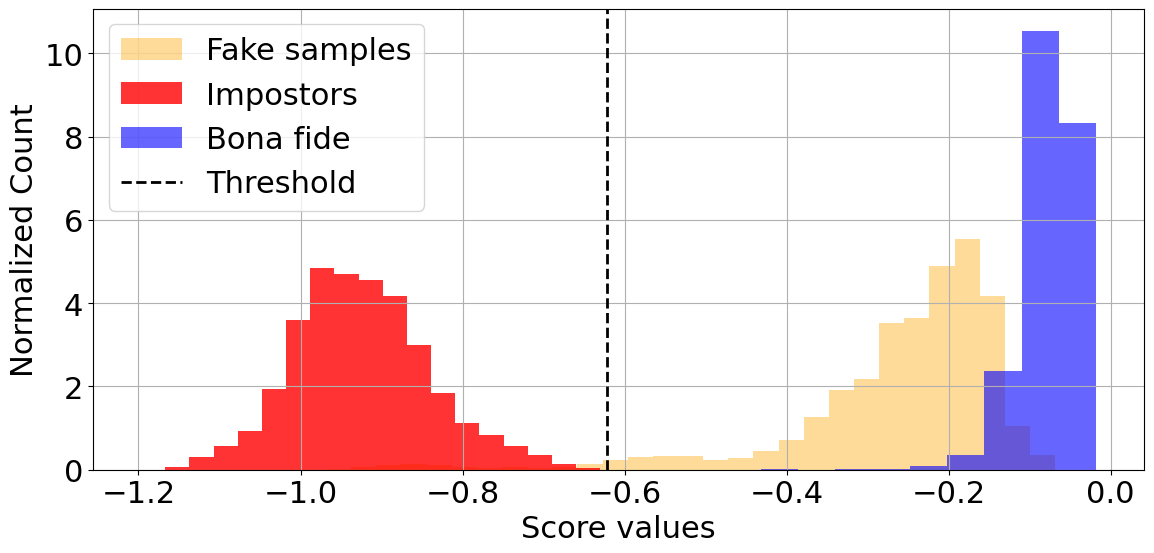}  
	\caption{Face deepfakes with 256px resolution model}
	\label{fig:vulnswan:a}
\end{subfigure}
\begin{subfigure}{0.33\textwidth}
	\includegraphics[width=\linewidth]{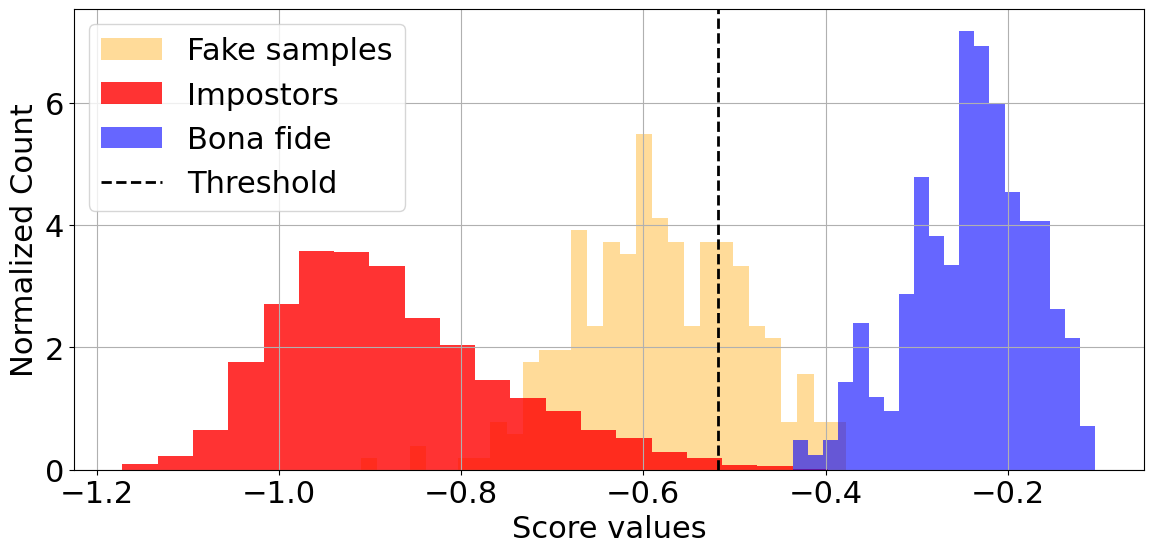}  
	\caption{Voice deepfakes with  YourTTS}
	\label{fig:vulnswan:b}
\end{subfigure}
\begin{subfigure}{0.33\textwidth}
	\includegraphics[width=\linewidth]{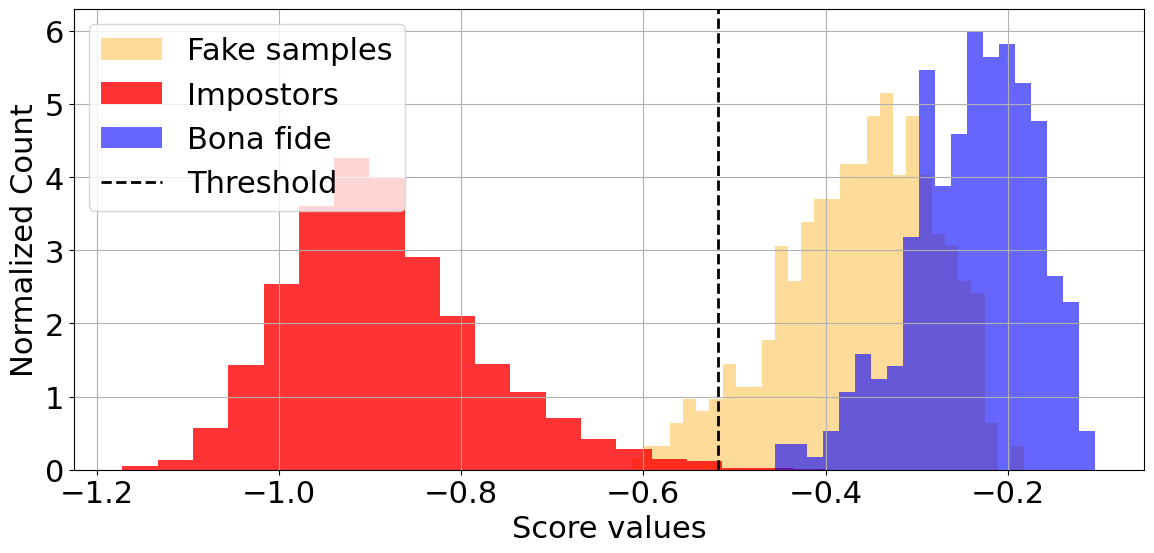}  
	\caption{Voice deepfakes with FreeVC}
	\label{fig:vulnswan:c}
\end{subfigure}
\caption{Score histograms of MobileFaceNet face and SpeechBrain speaker recognition models evaluated on the real videos (SWAN database) and a variant of generated deepfakes (SWAN-DF). The dotted vertical line marks the threshold computed on the validation set.}
\label{fig:vulnswan}
\end{figure*}

\section{Generative methods used in the database}
\label{sec:methods}

Our main goal is to create an audio-visual database of people speaking on camera where both video and audio channels are completely generated and which would look and sound as realistic as possible. We explored different methods for generating fake speech and fake faces and settled on the models by DeepFaceLab\footnotemark[6]~\cite{DeepFaceLab} for fake face swapping and FreeVC~\cite{li2022freevc} for voice conversion. In this section, we describe the methods and their variations that we used to generate audio-visual deepfakes for our SWAN-DF database. In addition, we describe the other speech generative methods that we could only use to create audio deepfakes resulted in LibriTTS-DF dataset.

\subsection{Video deepfakes}
\label{sec:video}

As a source of original videos, we selected $46$ different identities from a publicly available SWAN database~\cite{Swan2019}, which was recorded in $2019$. The videos in HD (resolution $720 \times 1280$)  include a person looking into iPhone or iPad Pro frontal camera and saying a set of phrases. From these $46$ identities, we manually matches $60$ pairs of people for face swapping process. In the selection process, we tried to match accessories, such as eye glasses, head and facial hair styles, skin colors, and genders. A well matched pair of faces typically leads to a visually more realistic deepfake. Since SWAN dataset has $16$ videos with sound per each person, swapping faces for $60$ pairs of people, results in $16 \times 60=960$ of the deepfakes per a given model architecture and a blending process (see frames extracted from the original and deepfake videos in Figure~\ref{fig:swandf} or view the videos on the demo page\footnotemark[11]).

To generate video deepfakes, we used a well known open source repository DeepFaceLab\footnotemark[6], which implemented two main GAN-based architectures the authors call DF and LIAE~\cite{DeepFaceLab}. We used three pretrained models provided by the DeepFaceLab community that can generate faces of $160 \times 160$ (DF architecture), $256 \times 256$ (LIAE architecture), and $320 \times 320$ (LIAE architecture) resolutions. The models are pretrained on the large datasets of `whole faces' (in DeepFaceLab terminology a facial area that includes chin and the half of a forehead) of several identities that allows models to learn the generic structure of a face and reduces the time required to tune the model to a specific pair of identities.

For each of the three model architectures and for each pair of identities, we tuned the pretrained model for $50K$ iterations, which resulted in about $4$ hours  for $160$px resolution model, $13$ hours for $256$px resolution, and $20$ hours for $320$px resolution on Tesla P40 GPU. Through the trial and error, we have selected some of the specific training parameters for each of the model and we provide these parameters in our open source packages\footnotemark[10]. For the model of $160$px resolution, we trained three different variants, including i) training face together with its face mask and with color correction on, ii) no masked training and no color correction, and iii) with mask training but no color correction. For the other two resolutions, we only trained model that included mask training and had no color correction switched on. In total, we tuned five different types of models for each of the $60$ swap-pairs.

Arguably, an important part of what constitutes a deepfake and makes it different from the real image, from a forensics point of view, is the blending technique that was used to place-in the generated face of a target into the original video frame of a source (see Figure~\ref{fig:swandf}). A DeepFaceLab GAN model typically generates a square image, in addition, it also learns mask of a face. This mask is used to cut off the face from the generated image and replace the source face with it. During the replacement process, the smoothing, blurring, warping, and color correction techniques can be applied to make the generated face look naturally fitting into the destination frame. These techniques, which we refer to as \textit{blending}, change the appearance of the resulted frame, introduce some unique distortions, and therefore impact the methods trained to detect the deepfakes. One could argue that the existing deepfake detection methods detect mostly the residues from the blending techniques rather than the patterns left by the GANs used to generate faces~\cite{Efros2020}. Therefore, to offset the lack of the variation of blending techniques in the existing deepfake datasets, we have used more than $20$ variants of deepfakes using different sets of blending parameters for the five models that we have trained for each swap pair. 
Each variation results in a differently looking deepfake face and we believe such variability in the dataset will be useful for the research community.

For ethical reasons, we selected not to publish open source code that makes it easier to create deepfakes, besides what is already available in DeepFaceLab repository\footnotemark[6], and also not to publish our trained models. We do however provide all of the parameters for training and blending that we have used in the process\footnotemark[10].



\subsection{Audio deepfakes}
\label{sec:audio}

We generated speech deepfakes using four voice conversion methods: YourTTS~\cite{YourTTS}, HiFiVC~\cite{HiFiVC}, DiffVC~\cite{DiffVC}, and FreeVC~\cite{li2022freevc} and two text to speech methods: Adaspeech~\cite{chen2021adaspeech} and TorToiSe TTS\footnotemark[7]. We did not use text to speech methods for our video deepfakes, since the speech they produce is not synchronized with the lip movements in the video. There are efforts to correct this issue by using an additional model, e.g., Wav2Lip\footnote{\scriptsize \url{https://github.com/Rudrabha/Wav2Lip}}~\cite{Wav2Lip2020}, to synchronize lips in video with a given speech, but they suffer from many visible artifacts and often produce unconvincing results. One notable effort that used the combination of a TTS method and Wav2Lip is the FakeAVCeleb~\cite{khalid2021fakeavceleb} audio-visual dataset, but the quality of the resulting videos is questionable. Some commercial systems use text to speech and then speech to video approaches to generate AI assistants, notably Synthesia\footnote{\scriptsize \url{https://www.synthesia.io/tools/text-to-speech-video-maker}}, but the synchronization issue persists there as well. 

Therefore, we used voice conversion methods (YourTTS, HiFiVC, DiffVC, and FreeVC) to generate fake speech for the SWAN-DF dataset, but we used text to speech methods (Adaspeech and TorToiSe TTS) with only one YourTTS voice conversion to generate a separate dataset based on LibriTTS~\cite{libritts}. 
We used the test-clean subset of LibriTTS dataset with $39$ speakers to generate deepfakes. We took a random $30$ utterances from a speaker to either tune a model (Adaspeech) or compute speaker embeddings (TorToiSe or YourTTS). We then generated fake samples from the same utterances used for tuning.

\begin{figure*}[tb]
\centering
\begin{subfigure}{0.49\textwidth}
	\includegraphics[width=\linewidth]{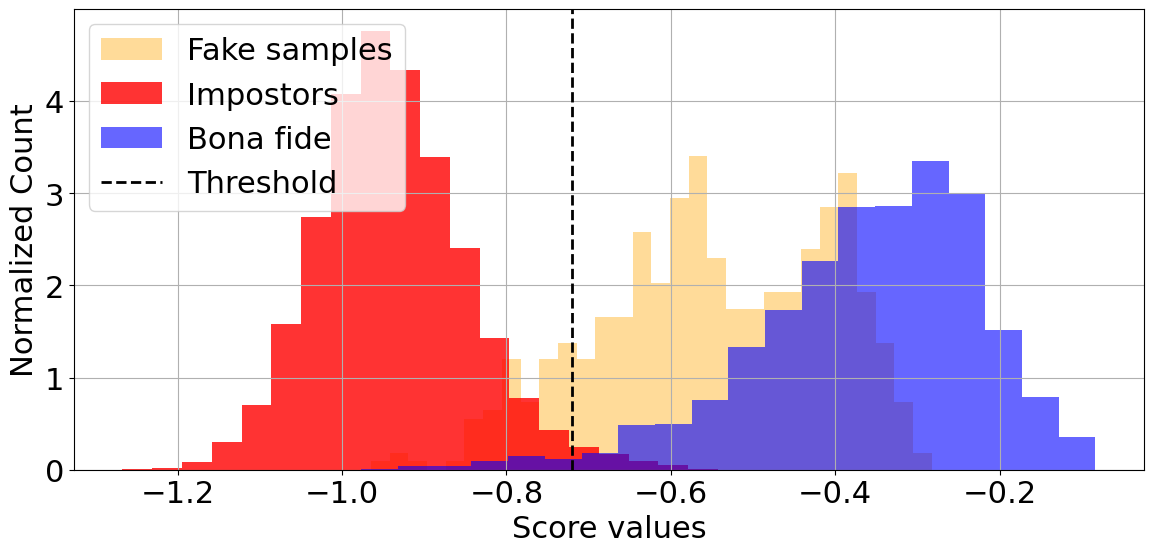}
	\caption{TorToiSe TTS}
	\label{fig:vulnlibri:a}
\end{subfigure}
\begin{subfigure}{0.49\textwidth}
	\includegraphics[width=\linewidth]{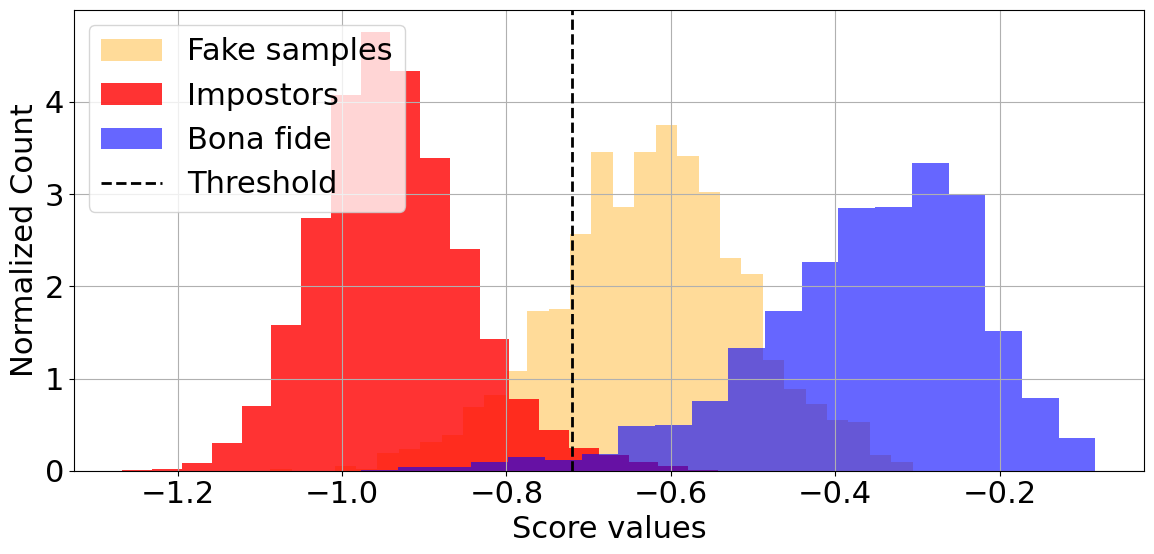}  
	\caption{YourTTS}
	\label{fig:vulnlibri:b}
\end{subfigure}
\caption{Vulnerability results of SpeechBrain speaker recognition model when evaluated on the original LibriTTS data and voice deepfakes generated by TorToiSe TTS and YourTTS models. The dotted vertical line marks the threshold computed on the validation set.}
\label{fig:vulnlibri}
\end{figure*}

A brief summary and the parameters of the speech deepfake methods used are as follows:
\begin{itemize}
	\item {\bf Adaspeech~\cite{chen2021adaspeech}:} a text to speech model specifically designed to be adapted to a custom voice and acoustic conditions. This model was further modified by us to allow adapting to a new voice with the aim of preserving the speaker identity. The model was pretrained to generate spectrograms from text on the popular VCTK corpus~\cite{VCTK} of speech from $109$ speakers reading $400$ sentences for $300K$ iterations. 
	This pretrained model was adapted (tuned) for each of $39$ speakers from LibriTTS~\cite{libritts} dataset using $30$ utterances of that speaker for $4K$ iterations. The adapted model is then used to generate the user-specific spectrograms from the provided text. HifiGan~\cite{hifigan} vocoder pretrained on LJSpeech~\cite{ljspeech17} database is then used to generate the final synthetic speech samples.

	\item {\bf TorToiSe TTS\footnotemark[7]:} a text to speech model is based on the autoregressive and diffusion encoders and is inspired by and very similar to DALLE~\cite{DALLE} for images. The author of this zero-shot generative model argues that it requires just a few seconds of reference speech to generate a high fidelity speech from any textual input. The model is pretrained on a set of speech databases, including a private one collected by the author, totaling about a million hours of speech. To generate our LibriTTS-DF database, we used $30$ utterances per speaker to compute the latent vectors of the model and then we produce the same but synthetic $30$ samples from the corresponding text. The preset `fast' of the model was used during the generation process.

	\item {\bf YourTTS~\cite{YourTTS}:} originally a text to speech model based on the end-to-end VITS~\cite{VITS} but with an addition of a separate speaker embedding (from a speaker recognition model~\cite{YourTTSspeakerencoder}) to encode speaker identity. The inclusion of speaker encoding allows the use of YourTTS in a zero-shot voice conversion manner by simply substituting the  embedding of one speaker with the embedding of another. We used the provided model pretrained on VCTK~\cite{VCTK} and LibriTTS~\cite{libritts} datasets. For our LibriTTS-DF dataset, we converted each speaker to a randomly chosen $5$ other  speakers, using only $10$ utterances for speaker encoding. For SWAN-DF dataset, we used all $16$ available utterances per each speaker for the encoding and converted the voices for the same pairs of speakers as in video deepfake swapping (see Section~\ref{sec:video} for details).

    \item {\bf HiFiVC~\cite{HiFiVC}:} a zero-shot many-to-many voice conversion system that relies on automated speech recognition (ASR) features, pitch tracking inspired by PPG-VC~\cite{PPGVC}, and their version of the waveform prediction model that extends HiFi GAN~\cite{hifigan}. We used provided model pretrained on VCTK dataset~\cite{VCTK} without any tuning. We used the same number of utterances as for YourTTS method to compute the speaker embeddings during the conversion. 
    
    \item {\bf DiffVC~\cite{DiffVC}:} another zero-shot many-to-many voice conversion method designed for the general case when source and target speakers do not belong to the training dataset. Since the authors emphasized that no tuning is required and for the provided model pretrained on LibriTTS dataset~\cite{libritts}, we used this method \emph{as is}. Please note that we used this method to generate deepfakes for SWAN-DF dataset, which is very different from LibriTTS. We used the same number of utterances as for YourTTS method to compute the speaker embeddings during the conversion.

	\item {\bf FreeVC~\cite{li2022freevc}:} an end-to-end model for voice conversion based on the approach proposed in VITS~\cite{VITS}. The model relies on WavLM features~\cite{WavLM} and a computationally heavy augmentation technique based on the resizing of spectrograms to several spectral bands, which allow to exclude noise from the data when learning speech characteristics. The provided model is pretrained on VCTK speech corpus~\cite{VCTK} and we have adapted it to the acoustic domain of our database by tuning the model on the mixture of subset from VCTK and data from SWAN dataset. We then convert voices for SWAN-DF using the same swap pairs as we did for video deepfakes (see Section~\ref{sec:video} for details). By using different tuning parameters of the FreeVC model, we produced $5$ different variants of data.

\end{itemize}

We obtain the complete audio-visual deepfakes by combining the videos produced by face swapping and the speech generated using voice conversion methods. We can match all face with all voice deepfakes, thus obtaining a very large set of videos where visual and voice channels are different. With $960$ videos/utterances in SWAN-DF for one deepfake variant, with more than $20$ variations of face swapping, and $8$ voice conversion variants, we can get more than $150K$ video combinations.

\section{Vulnerability assessment of deepfakes}
To evaluate how realistic the generated deepfakes are, we used the ECAPA-TDNN-based model from SpeechBrain\footnotemark[9], one of the best performing speaker recognition systems, and MobileFaceNet~\cite{MobileFaceNet2018}, which is one of the popular and practical face recognition models\footnotemark[10]. 
        
    \subsection{Evaluation protocol}
    \label{sec:adapt}

We assess the vulnerability of the speaker and face recognition to the deepfakes in the same way as the vulnerability of the biometric systems is assessed to the presentation attacks, as per the recommendation presented in the standard~\cite{iso_iampr}. Therefore, we report false match rate (FMR), which is similar to false positive rate (FPR), and false non-match rate (FNMR), which is similar to false negative rate (FNR), and impostor attack presentation match rate (IAPMR), which is the proportion of attacks that are incorrectly accepted as genuine samples by a biometric system (for details, see ISO/IEC 30107-3 standard~\cite{iso_iampr}). 

        
We split the deepfakes of our SWAN-DF and LibrtiTTS-DF databases into development and evaluation subsets roughly equal in size. We also ensured that the identities in the different subsets do not overlap. To compute the metrics, we define the threshold on the development set that corresponds to equal error rate (EER) computed on the real original data. We use this threshold when computing FMR and FNMR on the scores of the real data from the evaluation set, and also to compute IAPMR rate, when instead of the scores for zero-effort impostors, we use the scores corresponding to deepfakes. 



To demonstrate the accuracy of the selected recognition systems, we evaluated them on the real video data assuming no deepfake attacks are present. For both systems, we used the pretrained models provided by the respective repositories. We used only two real samples to enroll each identity and the rest of the samples from the identity were used for computing the error rates.  Using the EER threshold from the development set, SpeechBrain on the real audio from evaluation set resulted in very low $0.3$\% FMR and $0.0$\% FNMR values for the SWAN-DF dataset and $1.99$\% FMR and $1.82$\% FNMR values for the LibriTTS-DF dataset. Similarly, MobileFaceNet resulted in low $0.0$\% FMR and $0.05$\% FNMR values when computed on the real videos from evaluation set of SWAN-DF database. The red (zero-effort impostors) and blue-colored (bona fide samples) histograms in Figure~\ref{fig:vulnswan} and Figure~\ref{fig:vulnlibri} illustrate well the low FMR and FNMR values, since these histograms are clearly separated.

\begin{table}
\setlength\tabcolsep{6pt} 
\small
\centering
\begin{tabular}{llc}

{\bf Model, training params} & {\bf Blending Method} & {\bf IAPMR } \\ \toprule

160px, no mask & no blending & 96.27 \\
160px, mask training & seamless, mlk color & 95.22 \\
160px, mask training + color & overlay, no color & 96.43 \\
256px, mask training & params tuned & 96.78 \\
256px, mask training & overlay, no color & {\bf 97.36} \\
\bottomrule
\end{tabular}
\caption{Vulnerability of face recognition to selected variants of video deepfakes from SWAN-DF dataset.}
\label{tab:swanvideo}
\end{table}

\begin{table}
\centering
\begin{tabular}{llccc}

{\bf Approaches} &{\bf IAPMR } \\ \toprule

HiFiVC & 0.00 \\
DiffVC & 8.09 \\
YourTTS & 27.43 \\
FreeVC, not tuned & 15.44 \\
FreeVC, tuned 70K iterations & 92.59 \\
FreeVC, tuned 109K iterations & {\bf 94.21} \\
\bottomrule
\end{tabular}
\caption{Vulnerability of speaker recognition to selected voice conversion based deepfakes from SWAN-DF dataset.}
\label{tab:swanaudio}
\end{table}

\subsection{Vulnerability to SWAN-DF}

Table~\ref{tab:swanvideo} shows the IAPMR rates, computed for each video frame separately, using MobileFaceNet~\cite{MobileFaceNet2018} on the selected face deepfake variants we generated for SWAN-DF database. The high IAPMR rates in the table mean that more than $95$\% of the deepfake frames were recognized by MobileFaceNet as corresponding to the claimed real identity. This demonstrates that the evaluated face recognition model is highly vulnerable to the generated deepfake videos. Figure~\ref{fig:vulnswan:a} also illustrates this result by showing how the scores (yellow histogram) for deepfake variant generated using $256$px model (see the fourth row of Table~\ref{tab:swanvideo}) are next to the bona fide scores (blue histogram) and almost completely on the right side of the threshold (the dotted vertical line). 

Table~\ref{tab:swanaudio} shows similar IAPMR rates for the voice deepfakes generated with several voice conversion algorithms. The results in this table are quite different from those for the face deepfakes. The table shows that some voice conversion algorithms, notably HiFiVC and DiffVC, pose no threat to SpeechBrain speaker recognition system as it did not confuse the speech generated  by these algorithms with the claimed real identities. FreeVC without any tuning and zero-shot YourTTS show IAPMR rates above $15$\%, which are not that small, considering that FMR and FNMR for real voices are very close to zero (see Section~\ref{sec:adapt}). It appears that without tuning, the identities do not transfer well to the generated speech and do not pose a great threat to the recognition system. However, if we tune the model, such as the domain transfer we have done for FreeVC model, we can achieve the vulnerability level comparable to the face deepfakes with IAPMR rates higher than $92$\%. Also, the last two rows of the Table~\ref{tab:swanaudio} show that the longer tuning leads to the higher IAPMR rate. 

Figure~\ref{fig:vulnswan:b} and Figure~\ref{fig:vulnswan:c} illustrate the differences between zero-shot approaches like YourTTS and when we use the tuned model like FreeVC (tuned for $109$K iterations). These figures show that for the tuned model, the histogram of the deepfake scores shifts very near to the blue histogram of the bona fide scores.

\begin{table}
\centering
\begin{tabular}{llccc}

{\bf Approaches} &{\bf IAPMR } \\ \toprule

YourTTS & 80.06 \\
Adaspeech & 83.51 \\
TorToiSe & {\bf 86.61} \\
\bottomrule
\end{tabular}
\caption{Vulnerability of speaker recognition to voice conversion based deepfakes from LibriTTS-DF dataset.}
\label{tab:libri}
\end{table}

\subsection{Vulnerability to LibriTTS-DF}

Vulnerability evaluation on LibriTTS-DF database is interesting in a sense that it allows us to observe two important points i) the differences between voice deepfakes generated using text to speech and zero-shot voice conversion methods and ii) the different between using audio dataset recorded in the room with acoustic isolation like LibriTTS and the dataset recorded in a standard noisy office environment like SWAN. 

Table~\ref{tab:libri} shows IAPMR rates for zero-shot YourTTS, for zero-shot advanced diffusion-based TorToiSe TTS model and for Adaspeech, where the pretrained TTS model was tuned for each speaker. Surprisingly, the results show high IAPMR rates for all methods, with TorToiSe, despite being zero-shot model, being the most threatening to SpeechBrain speaker recognition system. Also note that from a personal subjective experience, TorToiSe TTS model produces the most realistic and pleasant to the ear audio utterances. Figure~\ref{fig:vulnlibri:a} shows how much more the scores for deepfakes generated by TorToiSe are close to the bona fide scores. Comparing with Figure~\ref{fig:vulnlibri:b}, it can be noted that, although IAPMR values are comparable, the scores for deepfakes of YourTTS model are noticeably nearer to the red distribution than the scores of TorToiSe deepfakes.

Comparing the result for YourTTS in Table~\ref{tab:libri} with the results in Table~\ref{tab:swanaudio}, we can notice that IAPMR value  $80.06$\% for LibriTTS-DF is significantly higher than  $27.43$\% value for SWAN-DF. Figure~\ref{fig:vulnswan:b} and Figure~\ref{fig:vulnlibri:b} illustrate this phenomena as well. In both cases, we used the same pretrained YourTTS model and similar approach on how we have generated voice conversion deepfakes. Therefore, the difference can be explained by the fact that the original SWAN dataset was recorded in an office environment using consumer smartphone and tablet, which had a significant effect on the resulted generative speech. Subjectively, the utterances produced by YourTTS for SWAN-DF dataset are much more noisy with a lot of metallic sounds compared to the same in LibriTTS-DF.



\section{Conclusion}

In this paper, we presented SWAN-DF database of high quality audio-visual deepfake that pose a high threat to state of the art speaker and face recognition systems. We also generated more than $20$ face swapping variants using the combination of different models and blending techniques. We  used $8$ different voice conversion methods from $5$ different models to generate voice deepfakes. Such large variation of the deepfakes in both audio and visual domains will allow researchers to develop multimodal, more robust, and generalizable methods for deepfakes detection. In addition to the database of audio-visual deepfakes, we released a LibriTTS-DF database of only voice deepfakes, which we generated using both voice conversion and text to speech method, including TorToiSe, one of the latest diffusion-based models. 

The deepfakes will remain to be a serious threat to the identity protection systems and a challenge for an automatic and human detection. Therefore, such high fidelity datasets like SWAN-DF  and LibriTTS-DF  play a key role in helping to overcome these challenges. 

\section*{Acknowledgements}
This work was funded by 
Swiss Center for Biometrics Research and Testing and NAST: Neural Architectures for Speech Technology, Swiss National Science Foundation grant\footnote{\scriptsize \url{https://data.snf.ch/grants/grant/185010}}. The authors would also like to thank Christophe Ecbert for the help with setting up face recognition system and Kevin Carta for the advice on which tuning parameters of DeepFaceLab deepfake generation models to use.

{\small
\bibliographystyle{ieee}
\bibliography{references}
}




\end{document}